# Class Algebra for Ontology Reasoning

Dan Buehrer and Lee Chee-Hwa
*Institute of Computer Science and Information Engineering*
*National Chung Cheng University*
*Chia Yi 621, Taiwan*
*http://www.cs.ccu.edu.tw/~dan*

## *Abstract*

*Class algebra provides a natural framework for sharing of ISA hierarchies between users that may be unaware of each other's definitions. This permits data from relational databases, object-oriented databases, and tagged XML documents to be unioned into one distributed ontology, sharable by all users without the need for prior negotiation or the development of a "standard" ontology for each field.*

*Moreover, class algebra produces a functional correspondence between a class's class algebraic definition (i.e. its "intent") and the set of all instances which satisfy the expression (i.e. its "extent"). The framework thus provides assistance in quickly locating examples and counterexamples of various definitions. This kind of information is very valuable when developing models of the real world, and serves as an invaluable tool assisting in the proof of theorems concerning these class algebra expressions.*

*Finally, the relative frequencies of objects in the ISA hierarchy can produce a useful Boolean algebra of probabilities. The probabilities can be used by traditional information-theoretic classification methodologies to obtain optimal ways of classifying objects in the database.*

## 1. Introduction

### 1.1 Ontology

The term 'ontology' is ambiguous. An ontology is a declarative representation of reality. An ontology is a concept description system in which all concepts are defined. Based on a declarative representation, concepts can stand for the set of their instances, can be used for reasoning, and can be used for communication.

### 1.2 Need for an ontology reasoning server

Ontologies are becoming increasingly important as agent-based technologies start to appear on the network (Adam Farquhar [13]). In the past, each company had its own format for its databases, and updates to the database were done via forms, which were different for each company. Now, as electronic commerce and agent negotiation become more common, it is necessary to develop a standard terminology for agents to use when they negotiate. It looks as though XML is becoming the standard language for sending signed online forms, but the definitions of the fields within the XML records still depend on the nature of the applications.

An ontology reasoning server is a useful tool for developing a common shared model and terminology. It acts as a database miner, searching for "interesting" subclasses and relations. It can create a decision tree which classifies the objects based on the most "meaningful" attribute at any node in the tree. It can also return a "description" of a set of objects. It can also make suggestions for Prolog-like rules, based on subset relationships between the extents of classes in its ISA hierarchy.

### 1.3 Class algebra

A class is seen mainly as an implicit set, and its definition involves mainly the class invariant, represented as a declarative logical expression. This class "invariant" gives the requirements for membership in the class. Class algebra then provides a common definition for the operations of union, intersection, difference, the dot operator, and generic relations and methods for these classes.

Definition 1:   A class algebra expression is recursively constructed from the following:
  A class name
  x+y   where x and y are class expressions (i.e. class union)
  x-y   where x and y are class expressions (i.e. class difference)
  x*y   where x and y are class expressions (i.e. class intersection)
  x . r   where x is a class expression and r is a binary relation
  x . inv( r )   where x is a class expression and inv( r ) is the inverse of a binary relation
  x where y   where x is a class expression and y is a logical "where" expression which will evaluate to "true", "false", or "unknown" for any input object "This".

Definition 2:   A class has the form <intent , extent>.
Here "extent" is the set of all objects in the database which return a "true" value for the intent.  The <intent> is a sorted disjunctive normal form (SDNF) [12].  The predicates in an intent generally provide constraints which specify what attributes and binary relations each object must have, along with the typing constraints for these attributes and relations.  It acts as the unique name of the class. We will describe how to form the SDNF in Chapter 3.

Definition 3:   The complete ISA lattice over a set of predicates P is a labeled lattice (i.e. a directed acyclic graph where any set of nodes have a unique least upper bound ancestor "union" node and greatest lower bound descendant "intersection" node).  Each node represents a class, and is labeled by its sorted disjunctive normal form.
  The complete ISA lattice has a unique top node labeled "any" whose extent contains all the objects in the database, and a unique bottom node labeled "empty" whose extent contains no objects, and whose sorted disjunctive normal form is "false".  A node whose intent has only one conjunct has sons formed by adding a predicate into the conjunct.  A node with an intent having a disjunction has sons where a predicate from one of the disjuncts has either been deleted or replaced by a more specific predicate.

Definition 4:   The "ISA hierarchy of a database state" is formed by possibly merging some of the neighboring nodes of the complete ISA lattice, thus getting a network.  The nodes which are merged are all "equivalent" for a given state of the database because the expressions for these nodes all produce the same "true", "false", or "unknown" values for all of the objects in the current state of the database.

Definition 5:   The union, intersection, and difference of classes are defined as
  $x + y := <\text{SDNF}(x.\text{intent} \lor y.\text{intent}), x.\text{extent} + y.\text{extent}>$
  $x * y := <\text{SDNF}(x.\text{intent} \,\&\, y.\text{intent}), x.\text{extent} * y.\text{extent}>$
  $x - y := <\text{SDNF}(x.\text{intent} \,\&\, {\sim}y.\text{intent}), x.\text{extent} - y.\text{extent}>$,
  where " $\lor$ " ,"&" and "~" are the Boolean operators "or", "and", and "not", respectively,

Class algebra provides a natural framework for comparing class definitions and letting several companies define a common ontology (i.e. an "ISA" hierarchy of classes which have attributes, binary relations, and methods).  It has several advantages over other object-oriented models.  For instance, Corba provides for object semantics, but does not involve a definition of an ISA hierarchy and inheritance.  ODMG 2.0 [4] includes a definition of multiple inheritance, but its bindings to various languages are unwieldy.  For instance, the Java binding does not currently define binary relations.  Models of languages such as C++, Java, and Smalltalk are quite complex, and disagree on such fundamental details as whether or not to include multiple inheritance, whether to use inheritance or delegation, how generic function calls are done, etc.  Object Navigator Notation [5] (ONN) is the most similar to class algebra, but the syntax becomes quite messy because of the inclusion of ternary relationships.  Also, ONN is currently just a notation used for object-oriented design, not for database queries.  Class algebra is a nice "simple" compromise which includes multiple inheritance, binary relations, and generic method calls with overriding.
  We have implemented a prototype ontology reasoning system in Java  This Java-based program allows users to define complex models by using class-algebra expressions.  These class-algebra expressions serve as a logical framework in which the set of all instances of objects that satisfy a class-algebra query is computable.  By using this environment, model designers can check out their definitions of concepts (i.e. classes) by seeing if the objects that they have defined are classified correctly.  Moreover, the containment relations between class extents can lead to the discovery of implication relationships between the class-algebra definitions.   The close connection between the class-algebra logical class definitions and the set of class instances is what makes this reasoning system so powerful.   The use of examples and counter-examples in artificial intelligence is well-known.  For example, Lenat's famous program AM [6] was able to discover the "interesting" concepts such as sum, product, prime numbers, and prime number pairs by looking at examples and counterexamples.

## 2. Related logical problems

### 2.1 Undecidability

Previous attempts at relating logic and sets have been hindered by the undecidability of most logical expressions.   This correspondence between a logical expression and the set of objects which it describes is not so clear for first-order logic.   Some first-order logic expressions are undecidable, which means that the set of objects which satisfy them is not recursively enumerable.   By limiting expressions to have one input variable, no constructor function symbols, and only temporary independent output variables which finally lead to "true", "false", or "unknown" output values, class algebra expressions become decidable.

The other necessary trick is to make use of an "active" complement operator, as in relational algebra, where the complement is only computed relative to the objects in a given state of a given database rather than relative to a whole Herbrand space of all possible functional expressions.   If we were to allow a successor function, for instance, we would immediately run into undecidable questions such as, "Is infinity even or odd?" for the axioms even(X)⇔~odd(X), even(X)⇔odd(succ(X)) and even(0).

### 2.2 "Non-Horn" axioms

Also, however, there are "non-Horn" axioms like p∨q.  If we want to use first-order logic to describe the fact that our current database is the only "correct" model, then we can add in the current state of the database as a set of facts.  However, we still need to rule out other models by adding "closed-world assumption" axioms, which make unprovable predicates false.   In the case of axioms like p∨q, this leads to a contradiction, since the axiom is not satisfiable in the presence of the axioms ~p and ~q, which were added by the closed-world assumption.

To overcome this problem, class algebra uses renaming of predicates, using the positive literal "false_p" instead of "~p".   Then the above axiom is transformed into the Horn clauses    p←false_q and q←false_p.  Like all Horn clauses, these clauses have a unique "least" model.   For this example, the least model is empty, containing neither p nor false_p, nor q nor false_q.   Of course, this is not a model for the original axioms, but it is a model for the transformed axioms.   This model corresponds to a model of the original axioms in "intuitionistic logic" [7].   We will discuss this in the next section.

## 3. Basic class algebra for index classes

Sometimes the terms "index class", "query class", or "view" are used to refer to an intentional class definition, where the class associated with an object changes as the object's attributes and relations change.   Class algebra expressions can be used to describe either index classes, as in this prototype system, or fixed class declaration constraints.

### 3.1 Basic form of where conditions

Any class algebra expression can be re-written as a class algebra expression of the form "any where <whereCondition>".   Class algebra "where" conditions use the ternary Boolean operators or (∨), and (&) and not (~).  The "where" conditions in the basic class algebra are only of the following forms:
```
     <wherecond> ::=   <wherecond> & <wherecond>   |   <wherecond> ∨ <wherecond>   |
          ~<wherecond>   |   '(' <wherecond> ')'   |   <basicpredicate>
     <basicpredicate> ::=   userDefinedAttrName in primitiveClass   |
          <attrexpr> containop setofPrimitiveValues   |   <attrexpr> relop Constant   |
          aggr(<attrexpr>) relop Constant
     <attexpr> ::=   <dottedBinaryRelation>.userDefinedAttrName   |   userDefinedAttrName
```

<primitiveClass> ::=   number   |   string
<relop> ::=   <   |   ~<   |   >   |   ~>   |   <=   |   ~<=   |   >=   |   ~>=   |   =   |   ~=
<containop> ::=   has   |   ~has   |   -has   |   ~-has   |   in   |   ~in   |   -in   |   ~-in

In these operators the "-" is a quasi-complement operator and the "~" is a true complement operator where c+~c=any.  The expressions involving "~" will succeed if any attribute or relation contained in the expression has a null value.

<aggr> ::=   cnt   |   sum   |   avg   |   std   |   min   |   max

The aggregate functions "cnt", "sum", "avg", "std", "min", and "max" can be used for numeric attributes, but only "cnt" can be used for strings.  The aggregate functions are the only "class" methods, referring to the whole class rather than to each individual "This" object of the class.  They can also be applied to attributes, which have list values.

<dottedBinaryRelation> ::=   <dottedBinaryRelation>.userDefinedRelationName   |   userDefinedRelationName

The sorted disjuctive normal form of a class algebra where expression is unique.  This was stated in [12] without a proof.  Before stating the proof here, we must first pay special attention to the subsumption of comparison predicates for the same attribute.  A simple subsumption check would not permit us to combine the predicates "age<30" and "age<40".  "Generalized subsumption" permits the use of some resolutions during the subsumption process.  Here, we need to use the rules (attr<Y ∨ attr<Z) & Y<Z ←→ attr<Z, and (attr>Y ∨ attr>Z) & Y<Z ←→ attr>Y.  In the computation of the disjunctive normal form, we can also use the following rule:

P & Y1<attr<Z1 ∨ P' & Y2<attr<Z2
←→ P & Y1<attr<Z1 ∨ P' & (min(Y1,Y2) < attr < max(Z1,Z2))
whenever P' & axioms |-- P.   For example, P & 30<age<50 ∨ P' & 40<age<60 will result in P & 30<age<50 ∨ P' & 30<age<60.

Here we are using the dual forms of standard resolution, so stronger conjunctions (with more literals) are subsumed by weaker conjunctions (with fewer literals).

Theorem 1:   The <whereCondition> of the sorted disjuctive normal form "any where <whereCondition>" of a class algebra expression obtained by the following steps is unique.

Step 1:  Replace "C+D where X" by "(C where X) ∨ (D where X)".   Replace "C*D where X" by "(C where X) & (D where X)".   Replace "C-D where X" by "(C where X) & ~(D where X)". Replace each nested class algebra expression "C.dottedRelns where X" by the term "inv(dottedRelns) in C & X".   If there are no dottedRelns, then use the term "This in C & X".   If there is no "where" term in the original expression, then use "This in C".

Step 2:   Repeatedly use the following rules to get a disjunctive form.
Rule 1:   x & (y ∨ z) --> x & y ∨ x & z   Distributivity of & over ∨
(y ∨ z) & x --> y & x ∨ z & x
Rule 2:   ~(x ∨ y) --> ~x & ~y       DeMorgan's Laws
~(x & y) --> ~x ∨ ~y
Rule 3:   ~ ~x --> x

Step 3:   Using the following rule to expand some predicates in a conjunct without changing the truth value of the expression.
P & Y1<attr<Z1 ∨ P' & Y2<attr<Z2   -->
P & Y1<attr<Z1 ∨ P' & (min(Y1,Y2) < attr < max(Z1,Z2)) if P' logically implies P.

Step 4:   Use the dual form of standard resolution to get all resolvents.
x&y ∨ ~x&z --> x&y ∨ ~x&z ∨ y&z
Also use generalized subsumption during the resolution process:
Subsumption:   x ∨ y --> x if y logically implies x
Notice that this involves the following rules as a special case:
Rule 1:   x & x --> x,   x ∨ x --> x
Rule 2:   x ∨ ~x --> true,   x & ~x --> false
Rule 3:   false & x --> false,   false ∨ x --> x
Rule 4:   true & x --> x,   true ∨ x --> true

Step 5:   Alphabetize each of the conjuncts in the conjunctive form by sorting the symbolic strings. Then alphabetize the conjuncts by treating each conjunct as a word.

Proof of theorem 1:  If any class algebra "where" expressions are logically equivalent, then they cover the same area in a Karnaugh map whose variables are the predicates of the where expression. We can use the dual form of standard resolution to get all the implicants which do not logically imply other implicants (i.e. all unsubsumed resolvents). It is well-known that these prime implicants are unique for the Karnaugh map. So all the equivalent expressions will produce the same set of prime implicants, and these prime implicants will subsume other stronger implicants. These prime implicants can be sorted by any well-defined ordering, such as their sorted alphabetic names. Thus all the equivalent expressions will produce the same sorted disjunctive normal form. This ensures that the sorted disjunctive normal form of a class algebra "where" expression is unique.

### 3.2 Dot operator and aggregate functions

The dot operator for binary relations results in a class whose extent is the union of the values of the relation for each input object. That is, relations are always "flattened" into a single set of objects, rather than being a set of sets of objects.

In contrast to a relational dot operator, a dot operator for attributes (i.e. lists of primitive-valued properties) results in the attribute lists being appended together rather than being unioned together. Appending is used so that the aggregate functions such as cnt, avg, and std will produce what we usually think of as the "correct" value.

### 3.3 Arguments

Class algebra "where" expressions have the operators "or", "and", and "not" connecting predicates whose arguments are all constant, except for one implicit argument "This" which is bound to the current object being tested for membership in this class. The computations of the arguments of a class algebra expression are all independent, and each argument results in a class with a specific sorted set of oids for the current state of the database. These results are finally given to the top-level predicates, which simply compute a Boolean value, and these Boolean values are combined via Boolean operations. It is quite clear that the operators <union, intersection, difference> are transformed into <or, and, and not> operators, respectively, when the expression is normalized.

### 3.4 Extension to user-defined predicates

Since an attribute value may be a list, the relop's are assumed to have an existential interpretation, so "classExpr.attr<=c" means that there exists at least one element of classExpr.attr which is less than or equal to c, and that "classExpr.attr" can't be null. It is very possible that the attribute list is both less than and greater than a given constant. Based on this existential interpretation, these simple class algebra predicates may be extended to any user-defined predicates or macros which always return a "true", "unknown", or "false" value for any given input object. Each object in a given database thus determines a unique value for each of the predicates, and thus a unique value for each class algebra expression, where we use ternary Boolean logic.

### 3.5 Complexity

If the domains and ranges of relations have a maximum cnt m, following a binary relation may require m unions of up to m objects each, but the result will be at most m objects. These unions can thus be done in time $O(m^2)$. By sorting the lists of oids, unions, intersections, and differences of sets of size m can be done in time $O(m)$. Aggregate functions can be calculated in $O(m)$ time. The reflexive-transitive closure operator may take up to $O(m^3)$ time if a simple algorithm is used. Since m is bounded by the number of oids, it may be considered to be constant, and the time taken for evaluating any constraint is thus $O(n)$, where n is the number of operators in the constraint. The main problem is when m gets very large, which may occur when taking complements. The number of objects satisfying the query "age~=200" may be very large. Notice, however, that when normalizing, the quasi-complement operator always results in positive predicates (e.g. age<>200), and the objects satisfying those predicates must at least have the given attributes (e.g. age) or relations in order to be included in the result. This usually helps to limit the sizes of the intermediate results.

### 3.6 Necessary and sufficient constraints

Class algebraic expressions can be used to define either the "necessary" constraints associated with data typing, or the "necessary and sufficient" constraints associated with database queries and index classes. For this paper, we are concerned mainly about the latter, since when building ontologies, a user wants to quickly retrieve all objects which satisfy a definition, or the objects which satisfy a superclass definition but fail to satisfy a subclass definition. This depends on the current values of the attributes and relations for those objects, not on their declared types. However, it is also helpful to also know which constraints are due to the declarations of classes and relations. The prototype system goes either way; either from the class and relations to their constraints and the objects satisfying those constraints, or from the selected objects to the constraints which describe them. In the latter case, an index class name may be assigned to these constraints.

### 3.7 Upper bound and lower bound

In reasoning, axioms and class algebra expressions are first normalized, as in the dual of resolution theorem proving, obtaining a disjunction of conjuncts of literals. Then for each literal of each conjunct, the complemented literal is made the "head" of a Prolog-like clause, and the other literals are put into the body of the clause. For instance, the conjunct (p & ~q & r) would result in the clauses "~p ← ~q & r", "q ← p & r", and "~r ← p & ~q", which are then renamed as Horn clauses "false_p ← false_q & r", "q ← p & r", and "false_r ← p & false_q". Such Horn clauses always have a unique "minimal" or "initial" model which is the set of all predicates that can be derived via forward chaining. This results in a set of predicates which are definitely true, a set which are definitely false (e.g. false_p means that p is false), some which are unknown (i.e. not provably true or false), and some which are both true and false (in which case there is no model for the original axioms). This gives us a rough set to bound all traditional models of the original axioms. Unfortunately, for an axiom like p∨q this procedure does not help us decide whether to put p or q in the model, but it simply gives us a lower bound in which neither p nor q is included, as well as an upper bound where they are both included.

Another way to look at this approach is to say that intuitionistic logic , on which probabilistic class algebra is based, provides a quick way of computing a lower bound of true classical models. The nodes in its network record all forward-chaining proofs for a given Horn set of clauses. By renaming the negated predicates (e.g. using false_p instead of ~p), we obtained a Horn set of clauses whose unique minimal model was the same as the union of the nodes in the network produced by intuitionistic logic (i.e. the set of all unit consequents of the axioms). This minimal model lists the predicates and the values of "This" for which the predicate is provably true or false. The "true" predicates give us a lower bound on all classical models, and the complement of the "false" predicates gives us an upper bound on all classical models of the original, untransformed axioms. In Chapter 4 we will extend this to an interval Boolean logic. For this logic, p and q both have values which are intervals [0,1], meaning that their values can be anywhere from false to true. If new evidence showed that false_p had probability at least 0.3, then q would also have probability at least 0.3 because of the rule q ← false_p. So Pr(p)= [0, 0.7] and Pr(q)= [0.3,1] in probabilistic class algebra.

### 3.8 Equivalence and implication

Two classes are "logically equivalent" if their intents are identical. Two classes are "equivalent for the database" if their extents are equal. Equivalence for the database is a necessary, but not sufficient, condition for logical equivalence.

A class D "logically implies" a superclass E if the intent of E can be derived from the intent of D by the operations of standard propositional logic. Since the intent is the class membership function, it is clear that if E can be derived from D, then the extent of D will be a subset of the extent of E, and D will be a subclass of E. Also, D's extent being a subclass of E's extent is a good indicator that it might be possible to derive E's intent from D's intent. In attempting this proof by backward chaining from E's intent, subgoals can be immediately dismissed if their extent contains objects which are not in E's extent.

### 3.9 ISA hierarchy

Equivalence classes can be arranged into a directed acyclic graph (i.e. a DAG), with smaller conjunctions near the top and longer conjunctions near the bottom. This is the ISA hierarchy which is inferred for a given state of the database. When completed with all sorted disjunctive normal forms, this would be a lattice structure, with any two nodes having a unique union node above them or intersection node below them. We are only concerned with a subgraph of this lattice, however. This network is determined by the values of objects in a particular state of a database. It is determined by merging the nodes of each equivalence class into one node, where the class algebra expressions all produce the same values for all of the database objects.

An ISA hierarchy node sometimes has more than one predicate which is added to its conjunct as you go down to one of its sons.  For example, the conjunct P might have a son "P&r&s" and one other son "P&~r".  If we know that for this database, either "P&r&s" or "P&~r" has objects in its extent and "P&r&~s" has no objects in its extent.  Then we can infer r→s for objects in P.  This kind of axiom may be true only for the current database state, but it is a useful suggestion of a lemma for an automatic theorem prover to try to prove.

Moreover, any set of objects has a unique set of predicates which produce the value "true" for all of those objects.  The conjunction of these predicates gives a unique conjunct node.  This new node is called the "description" of the set of objects, and it is simply the conjunction of all predicates returning "true" and the negations of all predicates returning "false" for all of the objects of the set. The description can be simplified by deleting all superclass predicates which are implied by the other subclass predicates.

Since the predicates contain no explicit variables, the class algebra expressions can be treated like propositional logic expressions.  All logically equivalent class algebra expressions can be assigned a unique normal form (i.e. a SDNF form) which can be used as the class's name.

When a user defines a new class, the intent is used as the class name.  The intent (and thus the class name) may be modified by adding an attribute constraint, by pushing the "Relation Constraint" button and adding a constraint on the number of objects in a relation, or by adding a union/intersection/difference typing constraint.  Each time that class constraints are modified, the SDNF form is recalculated.  If there is a class definition which exactly matches this SDNF form, then its name is displayed.  The name may be changed, but at most one class name can be assigned to each SDNF form.

## 4. Probability Boolean algebra

We can obtain counts of the number of objects satisfying any class algebra expression X or its quasi-complement  -X.  The counts can then be used to compute relative frequencies, which leads to an interval-based probability calculus that satisfies the laws of Boolean algebra.  With the addition of a number of "virtual objects" to satisfy user-provided constraints about relative frequencies, this probability algebra becomes a useful tool for deriving probability upper/lower bounds for each class algebra expression, although there are some limitations on the kinds of probability constraints that can be used.

We add a number of virtual objects to satisfy user-provided constraints about relative frequencies. Our algorithm uses the standard assumption of "minimal information change" in order to come up with a unique distribution which satisfies the probability constraints.   It says that the expression $\Delta\Sigma$-$Pr(C_i)\log(Pr(C_i))$ where $C_i$ is the defined class and $Pr(C_i)$ is the probability of class $C_i$, is minimal after adding the needed virtual objects into existing classes.  Adding virtual objects simply involves increasing a class variable "virtual_count", where each virtual count has a unique object identifier. The object identifiers are what permit our system to prevent objects from being counted twice in union classes, and this is what allows our system to satisfy the laws of probability such as $Pr(X+Y) = Pr(X) + Pr(Y) - Pr(X*Y)$.

### 4.1 Probability of a class algebraic expression

The probability of a class algebraic expression for a given state of a database is defined to be the relative frequency of the number of oids for which the expression is true divided by the total number of oids in the database.  The resulting probability algebra is also a Boolean algebra, where, for example, $Pr(p \lor any - p) = 1$, $Pr(false\_p \lor any - false\_p) = 1$, $Pr(p - p) = 0$, and $Pr(p \lor q) = Pr(p) + Pr(q) – Pr(p \& q)$.  The initial model of the transformed axioms satisfies the rules of Boolean algebra, even when predicates can take on fuzzy values between 0 and 1, as in fuzzy logic.

However, it must be understood that attr>5 is not the complement of attr<=5.  It is the complement of attr~>5, which has a universal interpretation, that either attr is undefined or all attr values are less than 5.  As far as the ontology reasoning system is concerned, the values of the predicates attr>5 and attr<=5 are completely unrelated.

### 4.2 Reasoning

How should we reason about independence, partitions, covers, subclasses, and other constraints that may influence the relative frequencies? The independence assumption, non-intersecting sets, partitions, and subset constraints can all be treated like constraints on the size of the intersection classes.

    indep(A, B) → Pr(A & B) = Pr(A) * Pr(B)
    nonoverlap(A, B) → Pr(A & B) = 0
    subset(A , B) → Pr(A & B) = Pr(A)

### 4.3 Probability constraints

Consider trying to satisfy a constraint $Pr(A \mid B) > 0.3$, where A and B are constraints. We will take this constraint to mean that at least 30% of the objects satisfying the constraint B must also satisfy the constraint A. Following we will consider adding constraints involving conditional and marginal probabilities, which we treat as relative frequencies.

How do we make the probability constraints to be satisfied in our ISA hierarchy?

(1). Consider trying to satisfy a constraint $Pr(A \mid B) > 0.3$. If the number of objects satisfying constraint A is already over 30% of the objects satisfying B, then nothing needs to be done. Otherwise, we simply add "virtual" objects into the intersection class A&B for which both A and B are true.

(2). How about a constraint like $Pr(A \mid B) < 0.6$? For the "<" or "<=" constraints, we add virtual objects into the class with the constraint B, if necessary.

(3). If there is a constraint $Pr(A \mid B) <> x$, then it can be replaced each by constraints, $Pr(A \mid B) > x$ ∨ $Pr(A \mid B) < x$. However, this kind of constraint is very difficult to satisfy. We do not permit this kind of constraint, or the complement (i.e. $Pr(A \mid B) = x$).

(4). For satisfying constraints of the form $Pr(A) < x$, we will simply add virtual objects to the "any" class, if necessary, and for satisfying constraints of the form $Pr(A) > x$, we will simply add virtual objects to the "A" class.

(5). We don't allow users to input the following type of probability constraints, since they can cause looping, and we cannot be decided whether the looping will terminate.

Type 1:     $Pr(A \mid B)$ op1 value1 & $Pr(B \mid A)$ op2 value2, where the op1 and op2 include ">" or ">=".

Type 2:     $Pr(X \mid A)$ op1 value1 & $Pr(Y \mid A)$ op2 value2, where the op1 and op2 include ">" or ">=", and X ≠ Y. This includes A = any.

Type 3:     $Pr(B \mid A)$ op1 value1 & $Pr(B \mid A)$ op2 value2, where the op1 includes ">" or ">=", and op2 includes "<" or "<=". This includes A = any.

Type 4:     If $Pr(A \mid B) >= c$ or $Pr(A \mid B) > c$ then label the edge e(A&B, B) with "⇕". If $Pr(A \mid B) <= c$ or If $Pr(A \mid B) < c$ then label the edge e(A & B, B) with "⇓". Following the directions indicated by "⇕" or "⇓" of the labeled edges, and the direction of ISA hierarchy edges going upward, if there is a cycle which passes through at least two disconnected labeled edges, then these constraints are not permitted.

We must make sure that we process from the bottom of the ISA hierarchy toward the top. We will then obtain a unique number of virtual objects that must be added in order to satisfy the probability constraints.

### 4.4 Algorithm for addition of virtual objects

The algorithm for bottom-up addition of virtual objects is as follows:

(1). For every conditional constraint, add edges into the ISA hierarchy labeled with (direction, operator, probability, modified). If $Pr(A \mid B) >= c$ or $Pr(A \mid B) > c$ then label the edge e(A&B, B) with (↑, >=, c, NO) or (↑, >, c, NO) respectively. If $Pr(A \mid B) <= c$ or $Pr(A \mid B) < c$ then label the edge e(A & B, B) with (↓, <=, c, NO) or (↓, <, c, NO), respectively. These include B=any.

(2). For every edge labeled with (direction, operator, probability, NO) whose corresponding constraint it is unsatisfied, call step (a) to process its starting node according to a bottom-up topological sort using the ISA hierarchy for the node corresponding to the sorted disjunctive normal forms of the starting node. Following are the substeps:

(a). If the corresponding constraint is $Pr(A \mid B) >= c$ or $Pr(A \mid B) > c$ then do the steps (b1)-(b2) for the starting node A&B. If the corresponding constraint is $Pr(A \mid B) <= c$ or $Pr(A \mid B) < c$ then do the step (c) for the starting node B. Then return to the caller.

(b1). Find the minimal value m which is the needed number of virtual objects of the nodes B and A&B. The formula for m is m = ceil( |B| * (c / (1 -c)) - |A & B| * (1 / (1-c))). Here the |A & B| and |B| indicate the old values of tot object counts of the nodes A&B and B respectively. If we cannot find the m, then output an error message and stop. Change the label to (↑, >=, c, YES) or

(↑, >, c, YES), respectively.  If ancestor classes of these nodes have sufficient virtual objects then move m virtual objects down to the node A&B, or move their own n virtual objects down to the node and add the m-n virtual objects into the node A&B.  In this step, every time there is a node X whose object count has increased ΔX, it should call step (d).

(b2).  For each edge e(X, A&B) labeled with (↑, operator, probability, Yes), follow the inverse direction of "↑" (i.e. from A&B to X) to check again and see X if the corresponding probability constraint is unsatisfied or not.  If we find the node X needs t virtual objects to satisfy the constraints, then move t virtual objects from node A&B down to the node X.  Repeat this procedure until meeting a satisfied constraint or until there are no further edges labeled with (↑, operator, probability, Yes).  In this step, every time there is a node Y whose object count had increased ΔY, it should call step (d).

We can prove that t is less than or equal to m such that there will be enough virtual objects in the node A&B to be moved down to the node X.  The |X| / |A & B| >= d was satisfied previously.  Now m is added to the virtual object count of the node A&B in order to satisfy the constraints corresponding to the corresponding edge (i.e. Pr(A | B) >= c).  We need to find the t which satisfies (|X| + t) / (|A & B| + m) >= d.  We can conclude that the minimal value of t can not exceed m.  The formula for t is t = ceil( d * |A & B| - |X| + d * m ).

(c).  Find the minimal count m of virtual objects needed for node B.  The formula for m is m =ceil( |A & B| / c - |B| ).  If we cannot find such an m, then output an error message and stop.  Change the label to (↓, <=, c, YES) or (↓, <, c, YES) respectively.  If ancestor classes have sufficient virtual objects then move m virtual objects down to the node B, or move their own n virtual objects down to the node B and add the m-n virtual objects into the node B.  In this step, every time there is a node X whose object count has increased ΔX, it should call step (d).

(d).  For every ancestor in the ISA hierarchy above node X, add ΔX into the tot object counts of the ancestor nodes, and push the edges connected to any node whose corresponding constraints are unsatisfied into a stack.  A class may be encountered many times following different paths.  It only can be added once.  Because every virtual object owns a unique oid.  Pop the stack to get one unsatisfied constraint and call step (a) until the stack is empty.  Then return to the caller.

### 4.5 Modeling belief intervals

Assume that there are two classes A and B with class algebra where expressions p and ~p respectively.  If we represent the lower bounds on their probabilities by t and f, respectively, then we can say that the where expression p has the belief interval [t, 1 - f].  If all expressions return either a true or a false value, with no unknown values, then the interval shrinks to one point, i.e. t = 1 - f.  In our system all user-defined predicates are ternary-valued, i.e. return either true, false or unknown, so that the relative frequencies of objects returning true or false produce an interval for where expressions.

## 5. What the prototype ontology reasoning system does

### 5.1 Define a class

The main user interface window is shown in Figure 1.  The user can first define some attribute names at the top of each column in the grid.  He can click on the attribute name to provide a constraint on that attribute.  He can then click "Hide" button on the top to hide the attribute; this result is similar to the projection operator in relation databases.  The constraints are listed in the box on the left.  When sufficient constraints have been entered, the user can provide a class name in the upper-lefthand input box.  The system can also add constraints involving restrictions on the number of objects in a relation., for each given input object.

### 5.2 Report on equivalences and implications

If you click on the "Implication Analysis" button, then a report window will display the logical equivalences, the classes which are equivalent for the database, the logical implications, and the implications for the database.  From this report, a user can sometimes find that the intent of one class is provable from the intent of another class when one extent is a subset of another extent.  Also we may find logically equivalent class algebra expressions when two classes are equivalent for the database.

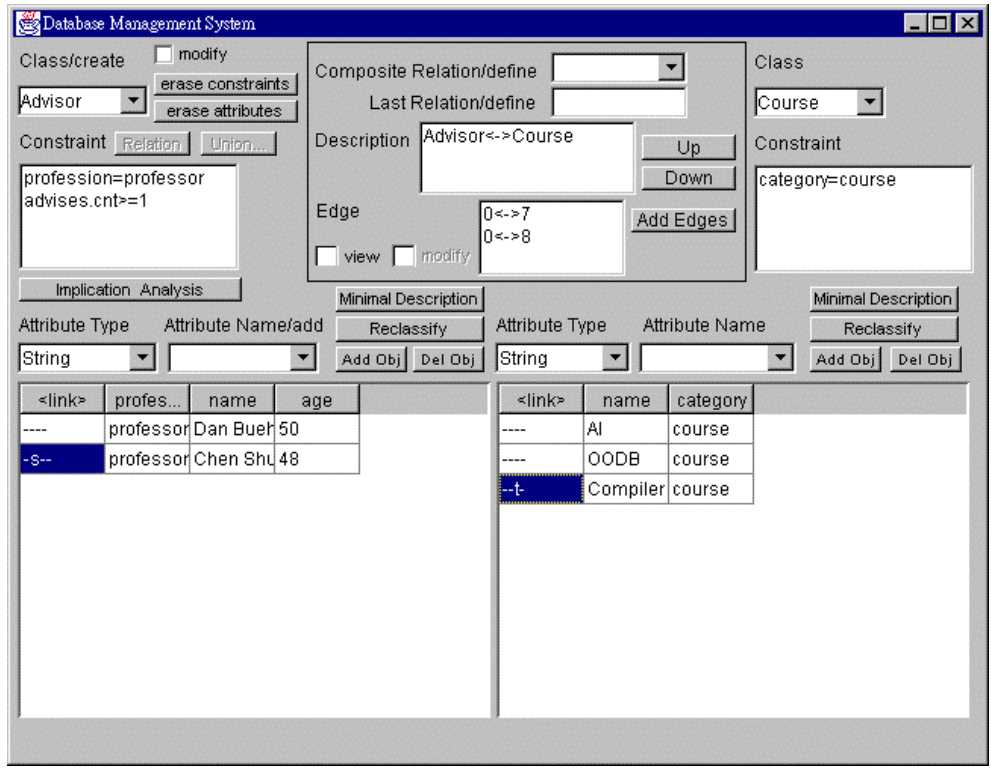

Figure 1. Main user interface window

### 5.3 Defining relations

Explicit binary relations can be defined by choosing a range class from the upper- righthand listbox. After selecting a "<link>" attribute on one row of the lefthand grid, you will see an "s" indicator representing the source of a binary link. Then selecting a "<link>" attribute for some rows on the right grid, you will see "t" indicators representing the targets of the binary link. Pushing the "Add Edges" button will add these links into the explicit binary relation. If two classes are chosen, but no edges are selected, then the relation which is being defined is assumed to be an implicit "class relation", relating all of the elements in one subclass to all of the elements of the other subclass. Composite relations are composed of dotted expressions involving explicit relations, implicit class relations, and other composite relations. You can use the "Down" button to move to next related pair of classes in the dot expression and then define the explicit relation or class relation between the pairs. At the last step you can define a relation name for the middle-top input box. At any time you can also give a name to the currently shown related pair of classes by providing a name for the input box below the composite relation name input box.

Methods are considered to be binary relations from the object "world" to itself. Each user has his own current "world" variable from which he can follow binary relations to access all other objects to which he has read/write/delete access privileges. Suppose that a superclass S has a subclass T which overrides some relation or method. Overriding of the relation or method is handled by moving the original definition down to the subclass S-T, and putting the new definition into subclass T.

### 5.4 Calculating the "description" of a set of objects

One powerful feature of the program is its ability to calculate the "description" of a set of objects (see Figure 2). By pushing the "Minimal Description" button, the user can also see what predicates are satisfied by all of the objects which have been selected in the grid. The predicates in the "description" are very often provable from the predicates in the constraint box, since they are true for all objects in the database. In the grid on the bottom of the report window, we also can find the probability that these objects belong to some class. This helps a user to find an approximate class if the data contains some noise.

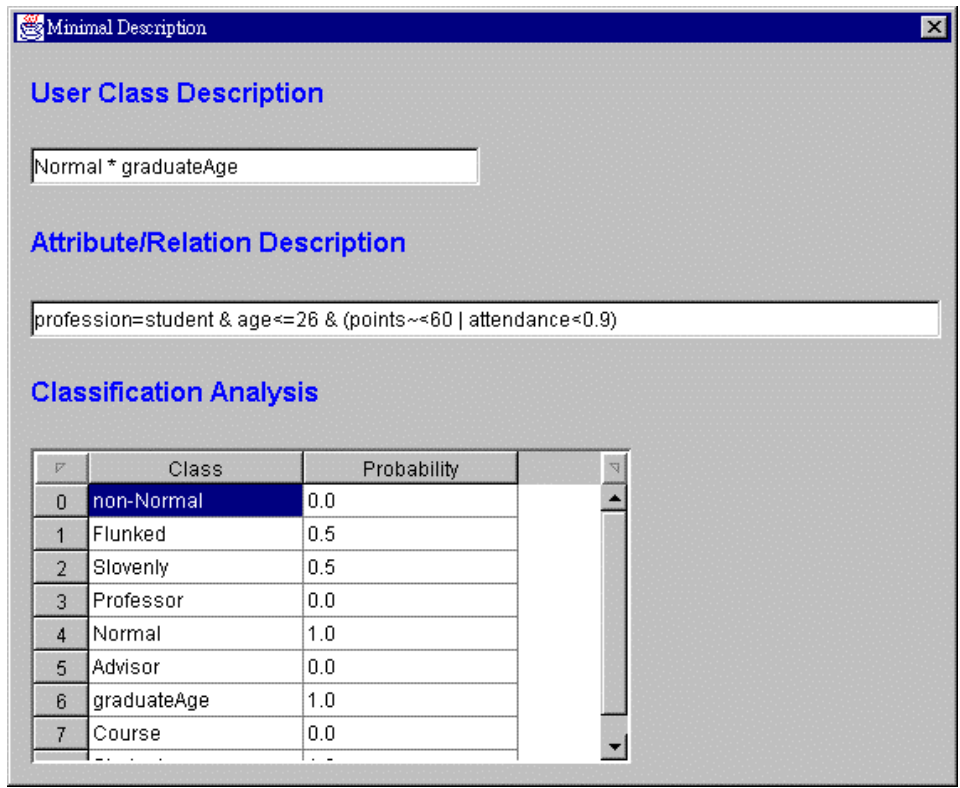

Figure 2. Report giving a "description" of a set of objects

**5.5 Helping the user to find relationships between sets of objects**

The user may also look for relationships among objects by clicking on one attribute and looking at a summary report which shows for each value of the attribute, the counts of the number of values and any aggregate functions of the other attributes. As in current Online Analytical Processing (OLAP), any dependencies between the attributes can quickly be seen from this report.

**5.6 Defining a class using other classes**

The Sorted Disjunctive Normal Forms of primitive classes in the system are simple conjunctions. They lack an "or" connector. The union of these primitive classes can provide the "or" connectors. Defining these classes can be done by clicking the "Union..." button and then selecting the needed classes.

**5.7 In the future**

Our Java-based ontology reasoning system is only a prototype. The program is run in a standalone environment now. In the future we will extend the system to support distributed ontology reasoning using JDBC and ODBC to link to relational databases, have an XML reader/writer interface to treat tables as human-readable documents, and also link to the class algebra API for our Java object-oriented database implementation. Moreover, in the future each attribute and relation will require its own B-tree index in order to be able to quickly identify all objects satisfying a particular class-algebra predicate.

## 6. Summary and conclusions

Class algebra is in between propositional logic and first-order logic. By disallowing variable interactions, and by using only Horn clauses, the undecidability problems of first-order logic can be overcome. In fact, for n objects, class algebra expressions can be computed in worst-case $O(n^3)$ time. Like propositional logic, all expressions have a sorted disjunctive normal form. For a finite number of predicates involving dotted expressions of limited length, there will only be a finite number of normal forms. A set of normal forms divides the objects of an object-oriented database into equivalence classes, where all of the objects in one equivalence class have the same values for the given normal forms. These equivalence classes are organized into an ISA hierarchy. This hierarchy is a subgraph of the complete lattice that corresponds to an intuitionistic logic interpretation of the Horn clauses which are the renamed version of the original class definitions. The intuitionistic logic lattice is good for any database, but the complete intuitionistic logic lattice structure is much too big to be practical. We can automatically select a subgraph of intuitionistic logic lattice which will classify input objects into "interesting" subclasses.

The class algebra described in this paper is a trivial kind of description logic; one with the inverse and complement operators for a finite domain (Calvanese [9]). Such description logics soon become undecidable after adding numeric quantifiers, such as "there exist at least 3" (Cadoli [10]). By eliminating variable interactions, class algebra remains decidable when adding constraints about the sizes of relations.

Because of the close, computable relationship between the logic expressions and the examples which satisfy it, class algebra is very helpful in defining models of reality. The examples and counterexamples of class definitions help users to understand the definitions and make corrections. Statistics on the number of objects in each subclass can also be very useful, and the relative frequencies can be used to define a Boolean algebra of probabilities. These probabilities can also be used to determine decision trees that contain the most information, as in the ID3 algorithm and its successor C4.5 [11].